\documentclass[10pt,letterpaper]{article}

\usepackage{cogsci}
\usepackage{pslatex}
\usepackage{apacite}
\usepackage{amsmath}
\usepackage{dsfont}
\usepackage{caption}
\usepackage{subcaption}
\usepackage{graphicx}
\usepackage{algorithm} 
\usepackage{algpseudocode} 
\usepackage{float}
\usepackage{multirow}
\cogscifinalcopy

\title{Experience-driven discovery of planning strategies}
\author{{\large \bf Ruiqi He (ruiqi.he@tuebingen.mpg.de)} \\
  Max Planck Institute for Intelligent Systems\\
  Tuebingen, Germany
  \AND {\large \bf Falk Lieder (falk.lieder@psych.ucla.edu)} \\
    UCLA Department of Psychology\\
  Los Angeles, United States}

\begin{document}

\maketitle

\begin{abstract}
One explanation for how people can plan efficiently despite limited cognitive resources is that we possess a set of adaptive planning strategies and know when and how to use them.
But how are these strategies acquired?
While previous research has studied how individuals learn to choose among existing strategies, little is known about the process of forming new planning strategies.
In this work, we propose that new planning strategies are discovered through metacognitive reinforcement learning.
To test this, we designed a novel experiment to investigate the discovery of new planning strategies.
We then present metacognitive reinforcement learning models and demonstrate their capability for strategy discovery as well as show that they provide a better explanation of human strategy discovery than alternative learning mechanisms.
However, when fitted to human data, these models exhibit a slower discovery rate than humans, leaving room for improvement.

\textbf{Keywords:} 
metacognitive learning, planning, strategy discovery, cognitive modeling, reinforcement learning
\end{abstract}

\section{Introduction}
From organizing dinner to drafting a summer holiday or mapping out a career, people engage in planning daily. 
Planning can be visualized as a decision tree that grows rapidly with more actions and steps to consider. 
While computers can rely on increasing computational power to solve such problems, human cognitive resources are limited. 
Yet, people often excel in complex domains where even advanced computers struggle, raising the question: how do people plan so efficiently despite cognitive limitations?
One explanation is that people use clever strategies to facilitate planning \cite{callaway2022rational}, but how do these strategies originate? 
While previous research shows how people learn to choose between pre-existing strategies from a mental toolbox through reinforcement \cite{rieskamp2006ssl}, little is known about how strategies enter this toolbox in the first place. 
Developmental psychology research \cite{siegler1991microgenetic} found that children can develop new strategies for solving arithmetic problems suggesting that strategy discovery may be a source of these strategies. 
While we understand some aspects of strategy discovery in mental arithmetic, the mechanism behind the acquisition of planning strategies is largely unknown. 
Therefore, in this work we investigate the question of how people discover new planning strategies. 


We address this question with two primary contributions:
First, we introduce a novel planning task requiring participants to learn an unfamiliar strategy to measure whether and to what extent strategy discovery occurs. 
Second, we present a computational model for strategy discovery based on the concept of \textit{metacognitive reinforcement learning} \cite{callaway2018learning}, which demonstrated a superior explanation of how people learn to adapt their amount of planning \cite{HeJainLieder2021NIPS-Planning} and planning strategies \cite{HeJainLieder2021}. 
Using the data collected from the new task, we then compare metacognitive reinforcement learning (MCRL) against alternative learning mechanisms to evaluate its potential as a model of human strategy discovery.


In the following sections, we will first present our experiment and empirical results. 
We will then discuss the various learning mechanisms, the resulting models, and the procedures for model fitting and selection. 
Finally, we will conclude with the modeling results.

\section{Measuring strategy discovery}
To measure strategy discovery, we first conducted an experiment using a process-tracing paradigm called Mouselab-MDP \cite{Callaway2017} and then analyzed the collected data to examine whether and to which extent strategy discovery took place.

\subsection{Mouselab-MDP}
\begin{figure*}
    \centering
    \includegraphics[width=0.8\textwidth]{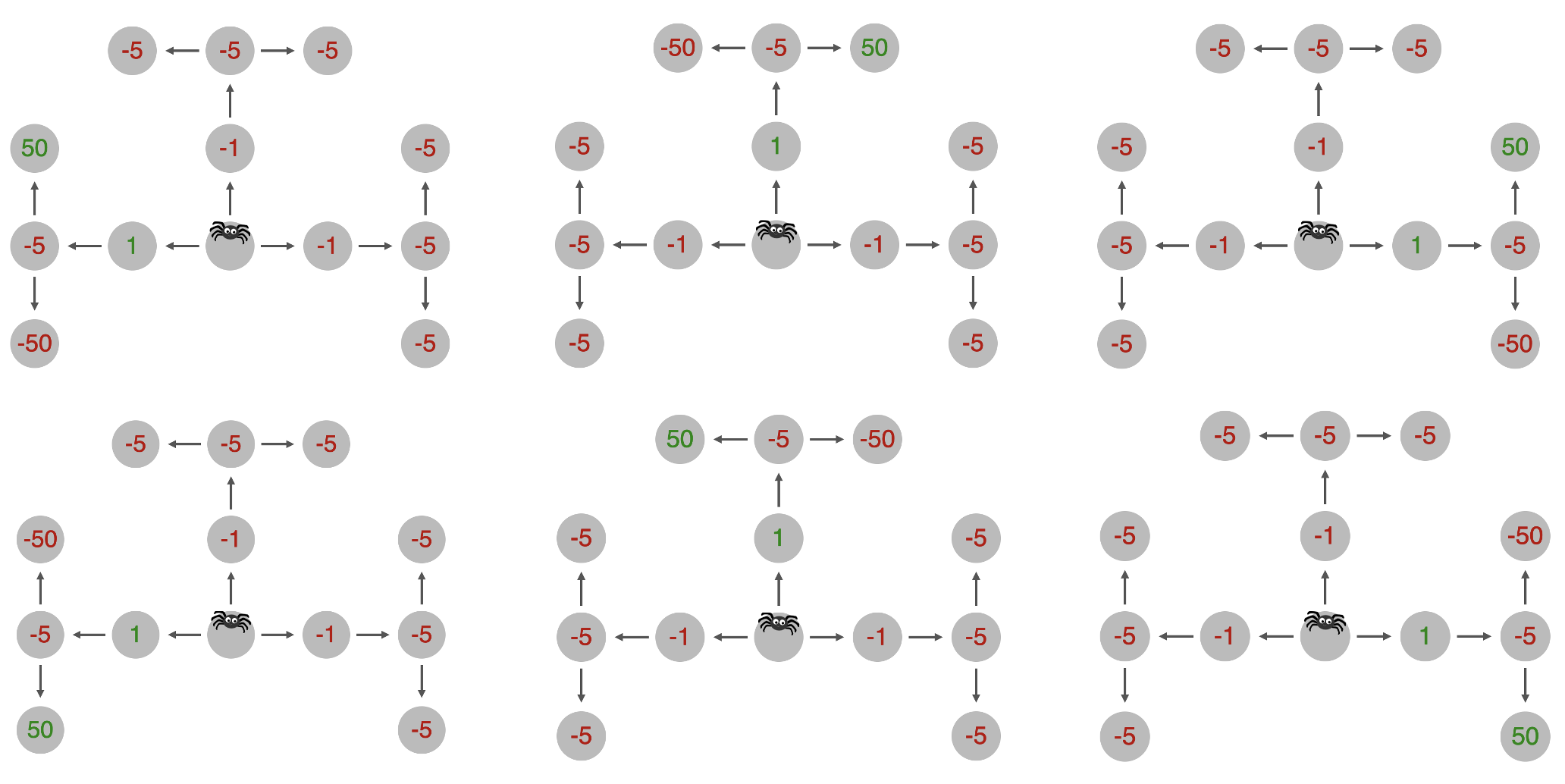}
    \caption{Each trial is randomly sampled from one of those six configurations.}
    \label{fig:strategydiscovery}
\end{figure*}

The Mouselab-MDP paradigm \cite{Callaway2017} extends the widely used Mouselab paradigm \cite{payne1976task} to address sequential decision-making problems. 
Essentially, the Mouselab-MDP presents participants with a planning task designed to render their information-gathering behavior highly indicative of their underlying planning strategies. 
Figure~\ref{fig:strategydiscovery} illustrates several example trials. 
Participants are asked to plan a spider's route through a maze with the goal to maximize their score. 
To track the planning process, the values associated with the nodes are initially hidden. 
Participants can click on a node to reveal its reward or loss.
This clicking behavior provides a manifestation of the participant's planning operations (i.e., the participant clicked on a node because they considered it in their planning process). 
The cognitive cost of this planning operation is externalized by imposing a fee for each click, encouraging participants to reveal information only when necessary. 
Therefore, the final score is the sum of the values of the gray nodes the participant chooses to traverse minus the fee for clicking. 
This score serves as a measure of resource-rationality \cite{lieder2020resource} that balances the expected utility with the cognitive cost of employing a particular strategy. 
The strategy that yields the highest score is referred to as the resource-rational (RR) strategy. 

\subsection{Experiment}

\paragraph{Materials}
We utilized the Mouselab-MDP paradigm to create an environment where the RR planning strategy was qualitatively different from any strategies participants might have previously known. 
In each trial, the environment could assume one of six possible configurations, as illustrated in Figure~\ref{fig:strategydiscovery}. 
Each configuration was designed so that exactly one branch began with a positive reward.
The outer nodes on the branch starting with an immediate positive reward always contained one node with a high positive reward (+50) and one node with a high negative reward (-50). 
The cost of revealing a node's hidden reward (click cost) increased with its distance from the starting point: 1 for the immediate nodes, 3 for the middle nodes, and 30 for the outer nodes. 
Consequently, the RR strategy involves first examining the immediate nodes until identifying or inferring the branch that began with a positive immediate reward. 
The RR strategy would then examine exactly one of the outer nodes on that branch.
If the value of that node was positive, the agent would choose the path leading to it; otherwise, it would choose the path leading to the other outer node on that branch.
\footnote{We confirmed that this is indeed the RR strategy for this environment by computing the environment's optimal metalevel policy using dynamic programming \cite{callaway2022rational}.}
Given the highly specialized nature of this strategy, we assumed that participants would not plausibly know it before the experiment.

\paragraph{Participants}
We recruited 420 participants through the platform CloudResearch, of whom 378 completed the experiment (average age 38.77 years; 240 females, 3 identifying as other). 
Recruitment was restricted to participants located in the United States who had completed more than 100 HITs and had an approval rate of over 80\%. 
Participants earned a bonus of 0.3 cents for each point in their final score. On average, they received a bonus of \$0.56 in addition to the base pay of \$4. 
The mean duration of the experiment was 25.8 minutes (median 22.8 minutes).
Participants who either did not finish the experiment or failed to click anything throughout all trials were excluded from further analysis. 
This exclusion criterion was designed to filter out inattentive participants.
After applying these criteria, 349 participants remained for analysis.

\paragraph{Procedures}
Each participant was given minimal instructions on how many trials they would face, how to reveal the nodes, and that each click would incur an unknown fee. 
They were also informed about the maximum and minimum values of all possible rewards and the score that the optimal strategy could achieve in each trial. 
After passing a quiz, they were asked to complete 120 trials of the planning task.

\subsection{Empirical results}
\begin{figure}[h!]
    \centering
    \includegraphics[width=0.8\linewidth]{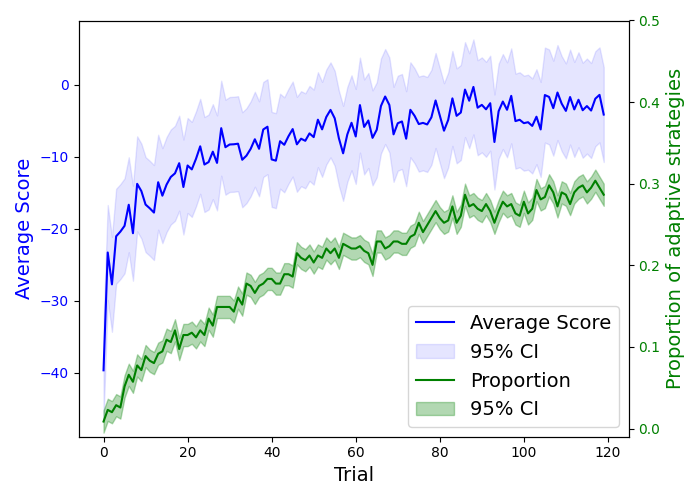}
    \caption{The average score and the proportion of the adaptive strategy plotted across 120 trials of all 349 participants.}
    \label{fig:scoreproportionplot}
\end{figure}

To investigate whether experience-driven strategy discovery is taking place, we first classified the participants' click sequences into two categories: adaptive strategies and other strategies. 
Adaptive strategies were defined according to the criteria outlined in the materials section. 
Using this classification, we plotted the proportion of adaptive strategies across trials and observed an increase from 0.79\% (CI: [0; 2.22\%]) in the first trial to 28.57\% (CI: [27.21\%; 29.93\%]) in the last trial (see the green line in Figure \ref{fig:scoreproportionplot}). 
A logistic regression analysis (\texttt{adaptive strategy $\sim$ intercept + trial}) confirms that this upward trend is significant, with a positive coefficient on the trial regressor, indicating an increasing proportion of adaptive strategies as participants gained more experience (see Table \ref{table:logisticregression}). 
This finding is further supported by a non-parametric Mann-Kendall test of trend ($S=6506, p<.001$).

\begin{table}[h!]
\centering
\begin{tabular}{cccc}
\hline
 & Coefficient & Std err & p-value \\ \hline
Intercept & -1.29 & 0.01 & \textless{}.001 \\
Trial & $4e^{-4}$ & $3e^{-5}$ & \textless{}.001 \\ \hline
\end{tabular}
\caption{Results of the logistic regression.}
\label{table:logisticregression}
\end{table}

Next to the significantly increasing proportion of adaptive strategy, a mixed linear regression on the score (\texttt{score $\sim$ intercept + trial}) also shows a significantly increasing trend (see Table \ref{table:linearregression} and the blue line in Figure \ref{fig:scoreproportionplot}). 

\begin{table}[h!]
\centering
\begin{tabular}{cccc}
\hline
 & Coefficient & Std err & p-value \\ \hline
Intercept & -9.39 & 0.57 & \textless{}.001 \\
Trial & 0.01 & 0.00 & \textless{}.001 \\ \hline
\end{tabular}
\caption{Results of the mixed linear regression.}
\label{table:linearregression}
\end{table}

Both results suggest that experience-driven strategy discovery is indeed taking place. 
However, strategy discovery appears to be a challenging task, as the proportion of participants who discovered the novel adaptive planning strategy increased only very slowly with only about 29\% of participants managing to learn it after 120 trials.

\section{Modeling strategy discovery}
Having established that experience-driven strategy discovery occurred, we now shift our focus to examining the underlying mechanism behind \textit{how} new planning strategies are discovered. 
We will first provide the necessary background, introduce the MCRL models, and demonstrate their ability for strategy discovery. 
Then to examine its alignment with human strategy discovery, we will introduce alternative learning mechanisms, outline the model fitting and selection procedures, and present the results.

\subsection{Reinforcement learning}
Reinforcement learning (RL) has proven to be a promising framework for modeling how people learn from interactions with the environment \cite{niv2009reinforcement}. 
Similar to human trial-and-error learning, a common type of RL involves predicting the potential reward $Q(s,a)$ for a specific action $a$ in a given state $s$, based on previous actions and the corresponding feedback $r$. The update rule for the Q-value is given by
$Q(s,a) \leftarrow Q(s,a) + \alpha \cdot \delta$
where $\alpha$ is the learning rate and $\delta$ is the reward prediction error which is the difference between actual and predicted reward.

\subsection{Metacognitive reinforcement learning (MCRL)}
While standard reinforcement learning focuses on learning policies for external actions, metacognitive learning operates on internal mental operations. 
To explore metacognitive learning, previous work \cite{krueger2017enhancing} utilized the MCRL framework to study the problem of deciding how to decide (meta-decision-making \cite{Boureau2015}) as a meta-level MDP \cite{hay2014selecting}: $M_{meta}=\left( \mathcal{B}, \mathcal{C} \cup \lbrace \bot \rbrace, T_{meta}, r_{meta} \right)$. 
In this framework, $b_t \in \mathcal{B}$ represents the mental belief state, $c_t\in \mathcal{C}$  denotes the mental computations including deliberation termination $\bot$. 
$T_{meta}$ describes the meta-level transitions between belief states, and $r_{meta}$ is the cost of computations as well as the expected return from terminating thinking and acting based on the current belief state. 
Solving a meta-level MDP optimally is often computationally challenging. 
However, the optimal solution can be approximated using MCRL by assuming that the brain estimates optimal meta-decision-making through reinforcement learning mechanisms \cite{Russell1991, callaway2018learning} that learn to approximate the optimal strategy directly \cite{HeJainLieder2021}. 

Before diving into the MCRL mechanisms, we will first introduce how the planning strategies are represented. 

\subsubsection{Strategy representation}
In the MCRL framework, planning strategies are represented by a set of features and corresponding feature weights. 
To represent the unique planning strategy of our planning task, we extended the feature set outlined by \citeA{Jain2021Computational}, who developed these features based on the Mouselab MDP paradigm, by adding seven new features that are essential to capture preferences and avoidance related to specific node levels, as well as one additional stopping criterion. 
Full details on these features can be found here: \url{https://osf.io/g3tzp}.
This extension results in a total of 63 features, which can be grouped into six broader categories believed to influence planning \cite{keramati2016adaptive, Daw2018}: Pavlovian features, model-free and heuristic features, mental-effort avoidance features, satisficing and stopping criteria features, model-based metareasoning features, and habitual features that reflect mental habits related to how frequently a node, branch, or level has been examined in the past and how many planning operations have been performed before. 
This strategy representation allows us to characterize an individual's learning trajectory as a series of combinations of features where the feature weights adjust over trials. 

After having described the features, we now turn to the MCRL models. 

\subsubsection{Gradient ascent through the strategy space (Reinforce)}
The metacognitive Reinforce model \cite{jain2019measuring}, based on the Reinforce algorithm by \citeA{Williams1992}, assumes that individuals adapt their planning strategies within the space of possible strategies. 
The strategies map features of belief states to values of cognitive operations, which are then transformed into probabilities of specific planning operations via a softmax function \cite{Williams1992}. 
Specifically, the strategy is parameterized by a weight vector $\mathbf{w}$, which is updated after each trial to reflect the effectiveness of different cognitive operations:
\begin{equation} \mathbf{w} \leftarrow \mathbf{w} + \alpha \cdot \sum_{t=1}^{O}\gamma^{t-1} \cdot r_{meta}(b_t,c_t) \cdot \nabla_\mathbf{w} \ln \pi_\mathbf{w}(c_t | b_t), \label{eq
} \end{equation} \label{eq1}
where $O$ represents the number of planning operations carried out during the trial, $b$ the belief state, $c$ the click under consideration, $\mathcal{C}_{b}$ the set of cognitive operations available in belief state $b$, $\alpha$ the learning rate, and $\gamma$ the discount factor. 
The learned weights, combined with the features described earlier, approximate the meta-level Q-values as follows:
\begin{equation} Q_{\text{meta}}(b_k,c_k) \approx \sum_{j=1}^{56} w_j \cdot f_j(b_k,c_k) \label{eq
} \end{equation}
These Q-values are then used to choose cognitive operations probabilistically, following the softmax function:
\begin{equation} P(c_k|b_k,Q_{\text{meta}}) \propto \exp(Q_{\text{meta}}(b_k,c_k) / \tau) \end{equation}
where $\tau$ is the inverse temperature that controls the balance between exploration and exploitation.
The free hyperparameters are $\alpha$, $\gamma$, and $\tau$ as well as the 63 initial feature weights.

We derived two variants of the Reinforce model: a hybrid variant that employs the full set of features including the model-based metareasoning features that capture the uncertainty of a node, as defined by the standard deviation and a purely model-free variant that utilizes a subset of the features by excluding the model-based metareasoning features, therefore positing that adaptation is driven purely by interactions with the environment (see first two rows in Table \ref{table:modeloverview}).

\subsubsection{Capability to represent and to discover the novel planning strategy}
\begin{figure}
    \centering
    \includegraphics[width=0.8\linewidth]{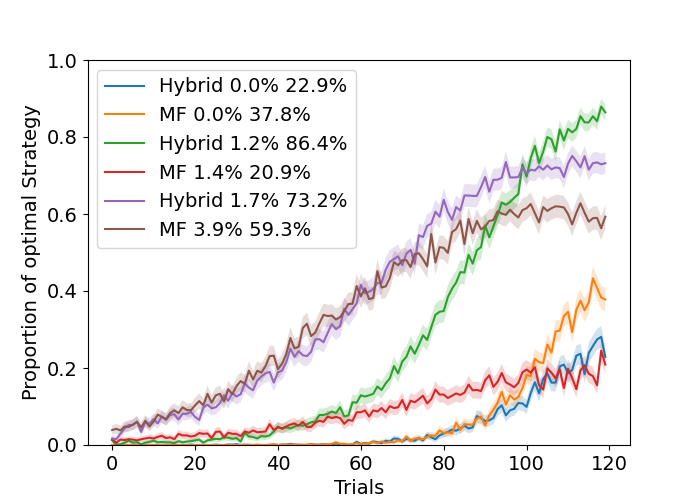}
    \caption{Simulations of the hybrid and model-free Reinforce models, each with 3 different sets of hyperparameters. Each curve shows the averaged performance across 1000 runs and the corresponding 95\% confidence intervals. Details on the hyperparameters of one of the simulations can be found here: \url{https://osf.io/g3tzp}.}
    \label{fig:reinforcesimulations}
\end{figure}

To evaluate our model’s capacity to represent and discover the adaptive strategy, we conducted model simulations using hyperparameters that were found by optimizing for the optimal click sequence. 
Figure \ref{fig:reinforcesimulations} shows that both the hybrid and model-free Reinforce models are capable of discovering the adaptive strategy from scratch (Mann Kendall test of trend for all simulations: $S>3442, p<.001$). 
While both the full and reduced set of features are sufficient to represent the adaptive strategy, it remains uncertain whether one model variant consistently outperforms the other.

\subsection{Alternative learning mechanisms}
Having demonstrated that our MCRL models are capable of strategy discovery, we now turn to examining how well our models can explain the human strategy discovery process. 
For this, we introduce two alternative learning mechanisms to compare against as well as a non-learning model. 

\subsubsection{Rational strategy selection learning}
Strategy selection learning \cite{rieskamp2006ssl}, posits that individuals possess a repertoire of cognitive strategies for problem-solving and that these strategies are reinforced through feedback. 
Similar to this idea, \citeA{LiederGriffiths2017} developed the model of rational strategy selection learning (RSSL), which frames the problem of deciding how to plan as an $n$-armed bandit problem, with each arm representing a different strategy.
In this model, each planning strategy is parameterized by the mean and variance of its expected return. 
Bayesian inference is performed on the expected return of each strategy, and strategies are selected using Thompson sampling from the posterior distribution of the expected return of the strategies. 
We will adopt the implementation described by \citeA{LiederGriffiths2017} and use the set of 79 strategies pre-defined by \citeA{Jain2021Computational}, which were extensively tested on the Mouselab-MDP.
Although the optimal strategy is not included in this set to reflect the constraints of working with a predefined set of strategies, the set contains a strategy that closely approximates the optimal one—specifically, a strategy that involves examining all immediate nodes and then one outer node at random.

\subsubsection{Mental habit formation}
According to this model, changes in planning strategy are not driven by value but are influenced by mental habits. 
It suggests that an individual's tendency to execute a particular planning operation increases with the number of times they have previously performed it regardless of the outcome, consistent with findings by \citeA{miller2019habits}. 
This approach involves using a weighted sum of frequency-based data, such as the number of prior clicks on the same node, branch, and level, within a softmax decision process. 
Technically, we implemented this model as a Reinforce model that uses the same set of features as the model-free variant but does not update the weights of the features. 
This model therefore adjusts its planning based on the frequency of previous planning operations from past experiences through the habitual features. 

\subsubsection{Non-learning model}
This model is implemented as a Reinforce model that does not perform any weight updates and does not use habitual features. 

In summary, we consider five different models: two MCRL models (hybrid and model-free Reinforce), the mental-habit, the RSSL, and the non-learning model. They all differ in the set of features used as well as whether and how the parameters are being updated (see overview in Table \ref{table:modeloverview}).

\begin{table}[]
\small
\begin{tabular}{ccc}
\hline
Model name & Updating parameters & Features \\ \hline
\begin{tabular}[c]{@{}c@{}}Hybrid \\ Reinforce\end{tabular} & Yes (updating weights) & \begin{tabular}[c]{@{}c@{}}Pavlovian, model-free, \\ satisficing, habitual, \\ model-based\end{tabular} \\
\begin{tabular}[c]{@{}c@{}}Model-free \\ Reinforce\end{tabular} & Yes (updating weights) & \begin{tabular}[c]{@{}c@{}}Pavlovian, model-free, \\ satisficing, habitual\end{tabular} \\
RSSL & \begin{tabular}[c]{@{}c@{}}Yes (updating mean \\ and variance \\ of each of the \\ 79 strategies)\end{tabular} & Not applicable \\
Mental habit & No & \begin{tabular}[c]{@{}c@{}}Same as model-free \\ Reinforce\end{tabular} \\
Non learning & No & \begin{tabular}[c]{@{}c@{}}Pavlovian, model-free, \\ satisficing\end{tabular} \\ \hline
\end{tabular}
\caption{Overview of the models, their features, and updating mechanism.}
\label{table:modeloverview}
\end{table}

\subsection{Model fitting and model selection}
After having introduced all the models, we now turn to comparing the MCRL mechanism against the alternatives. 
For this, all the models were fitted to each participant individually by maximizing the likelihood of their click sequences using Bayesian optimization \cite{bergstra2013making} for 60,000 iterations. 
The likelihood of a click sequence is calculated as the product of the likelihood of the individual clicks.
After fitting the models, we applied family-level Bayesian model selection (BMS) at three different aggregation levels. 
Specifically, we estimated the expected proportion of participants best described by each model family ($r$) and the exceedance probability ($\phi$), which indicates the likelihood that this proportion is significantly higher than that for any other model family using random effect Bayesian model selection \cite{rigoux2014bayesian}.
The first level of aggregation distinguishes between non-learning models and learning models. 
The second level compares the basic learning mechanisms: Reinforce, RSSL, and the mental habit model. 
The third level compares the two Reinforce variants: hybrid and model-free.
In addition to BMS, we used the Bayesian Information Criterion (BIC) \cite{schwarz1978estimating} to assess the relative fit of each model to individual participants' data.

\subsection{How well can the models explain human strategy discovery?}
BMS indicates that the majority of participants are better explained by a learning model (90\%) rather than a non-learning model (see Table \ref{table:bmsall}).
Among the learning models, the Reinforce models account for approximately 61\% of participants, outperforming the mental habit model (39\%) and RSSL ($<1\%$) with 100\% exceedance probability. 
Within the Reinforce variants, the hybrid Reinforce model explains 62\% of participants better than the model-free variant (38\%).
These observations suggest that while the MCRL models, especially the hybrid Reinforce, explains more participants better than the alternative learning mechanisms, there are individual differences in the learning mechanisms. 
Therefore, we next looked into the performance of the different participant groups. 

\begin{table}[h!]
\centering
\begin{tabular}{cccc}
\hline
Level & Model & $r$ & $\phi$ \\ \hline
\multirow{2}{*}{1} & Learning models & 89.65\% & 100.00\% \\
 & Non-learning model & 10.35\% & 0.00\% \\ \hline
\multirow{3}{*}{2} & Reinforce models & 60.89\% & 99.99\% \\
 & Mental habit model & 38.83\% & 0.01\% \\
 & RSSL & 0.29\% & 0.00\% \\ \hline
\multirow{2}{*}{3} & Hybrid Reinforce & 62.31\% & 100.00\% \\
 & Model-free Reinforce & 37.69\% & 0.00\% \\ \hline
\end{tabular}
\caption{BMS results where $r$ described the proportion of participants best explained by the model and $\phi$ the probability that this proportion is higher than the alternative.}
\label{table:bmsall}
\end{table}

\begin{figure}[h!]
    \centering
    \includegraphics[width=0.9\linewidth]{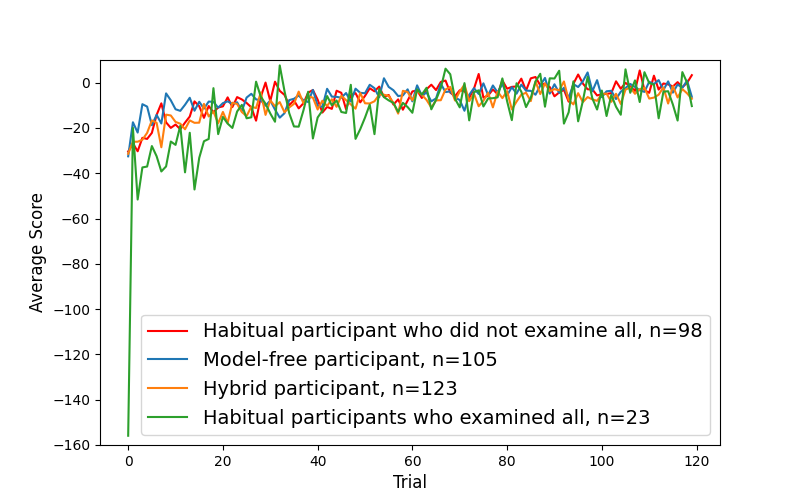}
    \caption{Average scores of the participants grouped by the model that has the lowest BIC. Participants who have been best explained by the mental habit model have been further subdivided into those who examined all the nodes in the first trial and those who did not.}
    \label{fig:groupedscores}
\end{figure}

When examining participants grouped by the model with the best goodness-of-fit (lowest BIC) an interesting pattern emerges. 
Participants best explained by the mental habit model (referred to as habitual participants), which does not learn but merely tends to repeat previous planning operations, performed better than those explained by the hybrid Reinforce mechanism (Mann-Whitney U test: mental habit vs. hybrid Reinforce $S=109740657.5, p<.001$; no significant difference noted when comparing habitual with the model-free Reinforce participants $S=90456709.5, p=.108$). 
The corresponding mean and standard deviations can be found in Table \ref{table:individualdifferences}. 
This counterintuitive observation can, however, be explained by some participants classified as habitual learners showing rapid improvement in the first 20 trials, followed by a flattened learning curve (see Figure \ref{fig:groupedscores}). 
Looking into the reason behind this explosive improvement reveals that 21 out of 121 habitual participants examined all the nodes early on (hence the extremely low score in the first trials),  thereby quickly learned the environment's structure, and then adopted the optimal strategy.
This type of insight-like learning was much less pronounced in other participant groups (15 out of 228 non-habitual participants). 
Therefore, these participants may be misclassified, as the mental habit model better captures their relatively flat learning curve following their abrupt change in behavior. 
This type of abrupt learning is currently not represented by any of the current models as they exhibit rather gradual improvements (see Figure \ref{fig:reinforcesimulations}). 

\begin{table}[]
\centering
\begin{tabular}{ccc}
\hline
 & Mean & Std \\ \hline
Mental Habit & -7.68 & 38.23 \\
Model-free Reinforce & -5.97 & 33.78 \\
Hybrid Reinforce & -9.04 & 37.99 \\ \hline
Kruskal Wallis test & \multicolumn{2}{c}{$S=37.76, p<.001$} \\ \hline
\end{tabular}
\caption{Mean and SD of the score of the participant groups best fitted by a model type and Kruskal-Wallis test testing the score for differences between all three samples.}
\label{table:individualdifferences}
\end{table}

\begin{figure}
    \centering
    \includegraphics[width=0.9\linewidth]{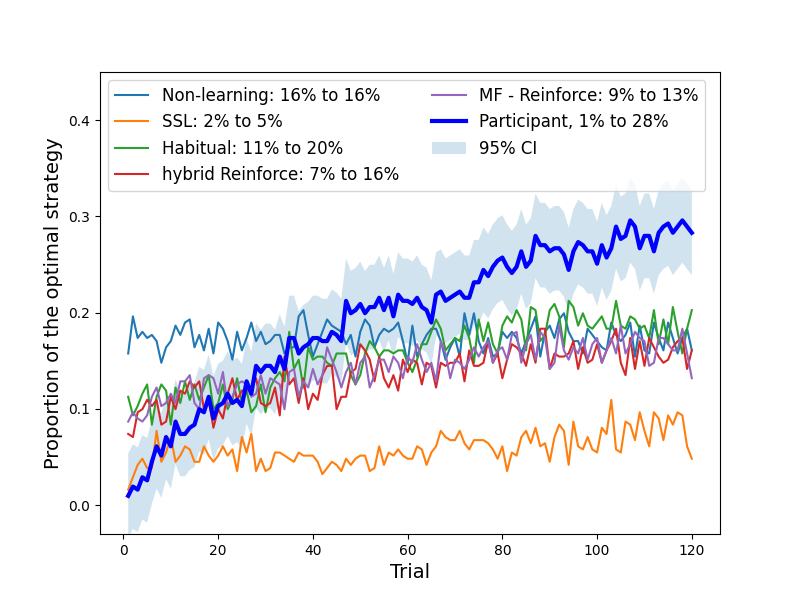}
    \caption{Model simulations using fitted hyperparameters of participants who did not examine all nodes at the first trial (n=311) and their proportion of the adaptive strategy at the first and last trial.}
    \label{fig:simulations}
\end{figure}

This limitation of the current MCRL models is reflected when simulating the discovery process using the models with hyperparameters fitted to the participants' clicking behavior. 
While the two Reinforce models, the mental habit model and RSSL, exhibit significant increasing trends in the proportion of adaptive strategies according to four Mann-Kendall tests (all $S>3024$ and all $p<.001$) as opposed to the non-learning model ($S=-115, p=.792$), participants discovered the new planning strategies at a different pace (see the increase in proportions in the legend of Figure \ref{fig:simulations}).
For each of the models, the slope was significantly lower (all $p<.001$, see \url{https://osf.io/g3tzp} for details on the regression analysis). 

\section{Discussion and further work}
This work investigated the underlying mechanisms of strategy discovery by designing an experimental paradigm that required learning a highly task-specific planning strategy that was unlikely part of participants' pre-existing mental toolbox. 
Empirical results show that while experience-driven strategy discovery was indeed taking place, it proved to be a challenging task, with only about 29\% of participants discovering the novel strategy after 120 trials. 
We then introduced the metacognitive reinforcement learning (MCRL) models and demonstrated their capability to discover new planning strategies. 
Fitting our models and alternative learning mechanisms to examine how well they could account for participants’ strategy discovery shows that 61\% of participants are better explained by the metacognitive Reinforce mechanism, while 39\% are better accounted for by a value-free mental habit model, highlighting i) that MCRL models are able to explain more participants better than the alternative learning mechanism and ii) individual differences in metacognitive learning. 
Moreover, while our models are in principle able to discover new planning strategies, fitting the models' hyperparameters on the participants' click sequences caused the models to discover adaptive strategies at a slower rate than human participants. 
We attribute this discrepancy to several factors. 
First, some participants seem to experience sudden, significant improvements in performance ('Eureka' moments) as well as engage in active learning - an aspect not captured by our current set of MCRL models. 
To address this, future work could explore additional mechanisms that incorporate active learning or insight-based models. 
Second, while our current feature set effectively represents the optimal strategy and captures a wide range of strategies, we cannot ensure that it encompasses all potential intermediate strategies participants might have adopted.

Overall, our work advances our understanding of metacognitive learning and how people discover adaptive planning strategies as well as contributes to the advances in artificial intelligence (AI) to reverse-engineer or even surpass human strategy discovery capacity.
While MCRL may not explain every aspect of this process, it is a key mechanism for uncovering strategies in complex tasks. 
We hope these insights enhance our understanding of human strategy discovery and inspire adaptive AI models across diverse domains.

\bibliographystyle{apacite}

\setlength{\bibleftmargin}{.125in}
\setlength{\bibindent}{-\bibleftmargin}

\bibliography{CogSci_Template}

\end{document}


\maketitle

This document provides supplementary details on implemented features, based on the implementation of \citet{Jain2021Computational}, the hyperparameters for two of the model simulations and details on the regression analysis. 

\section{Features} \label{appendix:featureslist}

\subsection{Existing features}
\textbf{Termination Constant }\\
The value of this feature is 1 for all clicks and 0 for the termination operation in all belief states. 
\subsubsection{Habitual features} 
\textbf{Ancestor count }\\
The value of this feature for a click in a given belief state is the number of ancestors of the considered node that have been observed in the current trial. 
\\
\\
\textbf{Depth Count }\\
The value of this feature for a click in a given belief state is the number of times that any node at the same depth as the considered node has been observed in the current trial. 
\\
\\
\textbf{Immediate successor count }\\
The value of this feature for a click in a given belief state is the number of children of the considered node that have been observed in the current trial. 
\\
\\
\textbf{Is the node a final outcome and has a positive ancestor?}\\
The value of this feature for a click in a given belief state is 1 if the considered node is a final outcome and it has an observed ancestor with a positive value in the current trial and 0 otherwise. 
\\
\\
\textbf{Is parent observed?}\\
The value of this feature for a click in a given belief state is 1 if the parent node of the considered node has been observed, and 0 otherwise. 
\\
\\
\textbf{Is the previous observed node a successor? }\\
The value of this feature for a click in a given belief state is 1 if the last observed node in the current trial is one of the successors of the considered node, and 0 otherwise. 
\\
\\
\textbf{Siblings Count }\\
The value of this feature for a click in a given belief state is the number of siblings of the considered node that have been observed in the current trial. 
\\
\\
\textbf{Squared Successor Count }\\
The value of this feature for a click in a given belief state is the square of the number of observed successors of the considered node for the current trial. 
\\
\\
\textbf{Successor Count }\\
The value of this feature for a click in a given belief state is the number of observed successors of the considered node for the current trial. 
\\
\\
\textbf{Depth }\\
The value of this feature for a click in a given belief state is the distance of the considered node from the starting position. 
\\
\\
\textbf{Feature 36: Observed height }\\
The value of this feature for a click in a given belief state is the length of the maximum observed path to a final outcome starting from the considered node. 
\\
\\
\textbf{Pavlovian features}
\\
\textbf{Best expected value}\\
The value of this feature for a click in a given belief state is the best expected return for a path in the current trial among all the paths that the considered node lies on. 
\\
\\
\textbf{Best largest value}\\
The value of this feature for a click in a given belief state is the maximum value observed among all the paths that the considered node lies on. 
\\
\\
\textbf{Does a successor node have a maximum value?}\\
The value of this feature for a click in a given belief state is 1 if any of the considered node’s observed successors in the current trial has a value which is the maximum possible value for the reward distribution, and 0 otherwise. 
\\
\\
\textbf{Value of the max expected return}\\
The value of this feature for all clicks in a given belief state is the maximum expected return from all paths in the current trial.
\\
\\
\textbf{Maximum value of an immediate successor}\\
The value of this feature for a click in a given belief state is the maximum value that has been observed among all the immediate successors of the considered node in the current trial. 
\\
\\
\textbf{Maximum value of a successor}\\
The value of this feature for a click in a given belief state is the maximum value that has been observed among all the successors of the considered node in the current trial. 
\\
\\
\textbf{Does the node lie on the most promising path?}\\
The value of this feature for a click in a given belief state is 1 if the considered node lies on the path with the highest expected return for the current trial, and 0 otherwise. 
\\
\\
\textbf{Value of the parent node}\\
The value of this feature for a click in a given belief state is the value of the considered node’s parent if the parent node has been observed, and 0 otherwise. 

\subsubsection{Model-based features}
\textbf{Current trial level standard deviation }\\
The value of this feature for a click in a given belief state is the uncertainty in the value of nodes at the same depth as the considered node as estimated for the current trial. 
\\
\\
\textbf{Max Uncertainty}\\
The value of this feature for a click in a given belief state is the is the maximum uncertainty in return for the current trial from all the paths that the considered node lies on. 
\\
\\
\textbf{Does the node lie on the second most promising path?}\\
The value of this feature for a click in a given belief state is 1 if the considered node lies on the path which has the second highest expected return for the current trial, and 0 otherwise. 
\\
\\
\textbf{Successor Uncertainty}\\
The value of this feature for a click in a given belief state is the total uncertainty in the values of all the successors of the considered node on the current trial. 
\\
\\
\textbf{Trial level standard deviation }\\
The value of this feature for a click is the uncertainty in the value of the considered node as estimated across all trials attempted so far by the agent. 
\\
\\
\textbf{Uncertainty}\\
The value of this feature for a click in a given belief state is the uncertainty in the value of the considered node. 

\subsubsection{Satisficing features}
\textbf{7 Satisficing features}\\
These seven features determine when the planning satisfices \cite{simon1956rational}. The value for these features is -1 for termination if the maximum expected return for the current trial is greater than certain satisficing thresholds. 
\\
\\
\textbf{Soft Satisficing}\\
The value of this feature for all clicks in a given belief state is the maxi- mum return that can be expected on the current trial from all paths that the considered node lies on. 
\\
\\
\textbf{Are all max paths observed?}\\
The value of this feature is -1 for all clicks and 0 for termination action in all belief states if all the paths path leading to a final outcome, which has the maximum value among the observed final outcomes, has been observed in the current trial and 0 otherwise. 
\\
\\
\textbf{Is a max path observed?}\\
The value of this feature is -1 for all clicks in all belief states if any path leading to the node, which has the maximum value possible for the reward distribution, has been observed in the current trial and 0 otherwise. 
\\
\\
\textbf{Is a positive node observed?}\\
The value of this feature is -1 for all clicks in all belief states if a node with a positive value has been observed in the current trial and 0 otherwise. 
\\
\\
\textbf{Is the previous observed node maximal?}\\
The value of this feature is -1 for all clicks if the last observed node in the current trial has the maximum value possible for the reward distribution and 0 otherwise. 
\\
\\
\textbf{Is a complete path observed?}\\
The value of this feature is -1 for all nodes in all belief states if at least one path has been completely observed from immediate outcome to final outcome, and 0 otherwise. 

\subsubsection{Pruning features}
\textbf{5 Pruning features}\\
These 5 features are designed to capture the idea of pruning branches \cite{Huys2012}. The value for these features is -1 for termination if the maximum expected return for the current trial is worse than certain pruning thresholds. 
\\
\\
\textbf{Is the previous observed node a successor and has negative value?}\\
The value of this feature for a click in a given belief state is 1 if the last observed node in the current trial is a child of the considered node and has a negative value, and 0 otherwise.

\subsection{Termination conditions}
\textbf{Immediate termination}\\
This feature returns 0 if no other node have been previously observed, and -1 otherwise. 
\\
\\
\textbf{Termination after first node}\\
This feature returns 0 if one node have been previously observed, and -1 otherwise. 
\\
\\
\textbf{Termination after positive root leaves}\\
This feature returns 0 if all leaf nodes have a path to an observed root node with a positive value, and -1 otherwise.

\subsection{New features}
\textbf{First level preference}\\
Preference for nodes that are one step away from the starting point. 
\\
\\
\textbf{Second level preference}\\
Preference for nodes that are two steps away from the starting point. 
\\
\\
\textbf{Third level preference}\\
Preference for nodes that are three steps away from the starting point. 
\\
\\
\textbf{First level avoidance}\\
Avoid nodes that are one step away from the starting point.
\\
\\
\textbf{Second level avoidance}\\
Avoid nodes that are two steps away from the starting point.
\\
\\
\textbf{Third level avoidance}\\
Avoid nodes that are three steps away from the starting point.
\\
\\
\textbf{Conditional termination}\\
Terminate after observing a positive inner node and one outer node on the same branch. 

\section{Hyperparameters}
\paragraph{Hybrid Reinforce}
\begin{itemize}
    \item Gamma: -0.011
    \item Inverse temperature: -0.600
    \item Learning rate: -9.580
    \item Initial weight on the feature all leaf nodes observed: -24.185
    \item Initial weight on the feature all roots observed: -0.899
    \item Initial weight on the feature ancestor count: 4.083
    \item Initial weight on the feature are max paths observed: 1.896
    \item Initial weight on the feature avoid first level: 7.710
    \item Initial weight on the feature avoid second level: 0.837
    \item Initial weight on the feature avoid third level: -2.231
    \item Initial weight on the feature best expected: -0.162
    \item Initial weight on the feature best largest: 0.310
    \item Initial weight on the feature branch count: -3.621
    \item Initial weight on the feature click count: -7.801
    \item Initial weight on the feature constant: 6.277
    \item Initial weight on the feature count observed node branch: 12.507
    \item Initial weight on the feature depth: -0.745
    \item Initial weight on the feature depth count: -9.795
    \item Initial weight on the feature first level: 1.058
    \item Initial weight on the feature get level observed std: 1.104
    \item Initial weight on the feature hp -24: 11.334
    \item Initial weight on the feature hp -48: -1.320
    \item Initial weight on the feature hp -8: 4.223
    \item Initial weight on the feature hp 0: -1.484
    \item Initial weight on the feature hs 0.0: 4.838
    \item Initial weight on the feature hs 16.0: -0.539
    \item Initial weight on the feature hs 24.0: 1.375
    \item Initial weight on the feature hs 32.0: 9.734
    \item Initial weight on the feature hs 40.0: 0.087
    \item Initial weight on the feature hs 48.0: 6.504
    \item Initial weight on the feature hs 8.0: -2.794
    \item Initial weight on the feature immediate successor count: -1.099
    \item Initial weight on the feature is leaf: 20.615
    \item Initial weight on the feature is max path observed: -5.304
    \item Initial weight on the feature is pos ancestor leaf: -0.406
    \item Initial weight on the feature is positive observed: 0.096
    \item Initial weight on the feature is previous max: 3.762
    \item Initial weight on the feature is previous successor negative: 0.678
    \item Initial weight on the feature is root: 0.405
    \item Initial weight on the feature is successor highest: -0.007
    \item Initial weight on the feature level count: -5.227
    \item Initial weight on the feature max expected return: 10.966
    \item Initial weight on the feature max immediate successor: 15.501
    \item Initial weight on the feature max successor: 2.349
    \item Initial weight on the feature max uncertainty: 0.210
    \item Initial weight on the feature most promising: 1.926
    \item Initial weight on the feature num clicks adaptive: 1.029
    \item Initial weight on the feature observed height: 0.734
    \item Initial weight on the feature parent observed: -17.009
    \item Initial weight on the feature parent value: -19.719
    \item Initial weight on the feature positive root leaves termination: 16.189
    \item Initial weight on the feature previous observed successor: -0.727
    \item Initial weight on the feature second level: -3.947
    \item Initial weight on the feature second most promising: -3.747
    \item Initial weight on the feature siblings count: -0.260
    \item Initial weight on the feature single path completion: -1.075
    \item Initial weight on the feature soft pruning: 8.520
    \item Initial weight on the feature soft satisficing: 6.448
    \item Initial weight on the feature sq successor count: 5.950
    \item Initial weight on the feature successor count: 13.358
    \item Initial weight on the feature successor uncertainty: -8.152
    \item Initial weight on the feature termination after observing positive inner and one outer: -19.781
    \item Initial weight on the feature termination constant: -15.968
    \item Initial weight on the feature third level: 11.377
    \item Initial weight on the feature trial level std: -7.485
    \item Initial weight on the feature uncertainty: -6.814
\end{itemize}

\paragraph{Model-free Reinforce}
\begin{itemize}
    \item Gamma: -0.038
    \item Inverse temperature: -0.963
    \item Learning rate: -10.220
    \item Initial weight on the feature all leaf nodes observed: -4.803
    \item Initial weight on the feature all roots observed: 13.337
    \item Initial weight on the feature ancestor count: 9.967
    \item Initial weight on the feature are max paths observed: 3.965
    \item Initial weight on the feature avoid second level: 18.046
    \item Initial weight on the feature avoid third level: -5.468
    \item Initial weight on the feature best expected: -0.967
    \item Initial weight on the feature best largest: 1.042
    \item Initial weight on the feature branch count: 2.481
    \item Initial weight on the feature click count: -12.353
    \item Initial weight on the feature constant: 0.385
    \item Initial weight on the feature count observed node branch: 10.712
    \item Initial weight on the feature depth: -0.228
    \item Initial weight on the feature depth count: 6.843
    \item Initial weight on the feature first level: -1.147
    \item Initial weight on the feature hp -24: 36.165
    \item Initial weight on the feature hp -48: 1.531
    \item Initial weight on the feature hp -8: 17.811
    \item Initial weight on the feature hp 0: 1.316
    \item Initial weight on the feature hs 0.0: 22.722
    \item Initial weight on the feature hs 16.0: -2.981
    \item Initial weight on the feature hs 24.0: 4.315
    \item Initial weight on the feature hs 32.0: -9.353
    \item Initial weight on the feature hs 40.0: 0.212
    \item Initial weight on the feature hs 48.0: 4.486
    \item Initial weight on the feature hs 8.0: -12.569
    \item Initial weight on the feature immediate successor count: -7.819
    \item Initial weight on the feature is leaf: 31.170
    \item Initial weight on the feature is max path observed: 0.571
    \item Initial weight on the feature is pos ancestor leaf: 0.134
    \item Initial weight on the feature is positive observed: 17.585
    \item Initial weight on the feature is previous max: -5.603
    \item Initial weight on the feature is previous successor negative: 2.420
    \item Initial weight on the feature is root: -4.010
    \item Initial weight on the feature is successor highest: -0.549
    \item Initial weight on the feature level count: 0.499
    \item Initial weight on the feature max expected return: 15.971
    \item Initial weight on the feature max immediate successor: -8.241
    \item Initial weight on the feature max successor: -0.333
    \item Initial weight on the feature most promising: -18.115
    \item Initial weight on the feature num clicks adaptive: -6.537
    \item Initial weight on the feature observed height: -2.520
    \item Initial weight on the feature parent observed: -4.105
    \item Initial weight on the feature parent value: -34.131
    \item Initial weight on the feature positive root leaves termination: 0.895
    \item Initial weight on the feature previous observed successor: -2.621
    \item Initial weight on the feature second level: 23.600
    \item Initial weight on the feature second most promising: -3.218
    \item Initial weight on the feature siblings count: 1.062
    \item Initial weight on the feature single path completion: -3.151
    \item Initial weight on the feature soft pruning: 24.239
    \item Initial weight on the feature soft satisficing: -9.083
    \item Initial weight on the feature sq successor count: -0.936
    \item Initial weight on the feature successor count: -9.804
    \item Initial weight on the feature termination after observing positive inner and one outer: 9.072
    \item Initial weight on the feature termination constant: -0.168
    \item Initial weight on the feature third level: -12.978
\end{itemize}

\section{Regression analysis on the slope of the models}
\begin{table}[h!]
\centering
\begin{tabular}{ccccccc}
\hline
Variable & Coefficient & std & t-value & p-value & 0.025 & 0.975 \\ \hline
Participant & 0.074 & 0.003 & 25.872 & \textless{}.001 & 0.068 & 0.080 \\
Habitual model & 0.037 & 0.004 & 9.103 & \textless{}.001 & 0.029 & 0.045 \\
MF Reinforce & 0.040 & 0.004 & 9.920 & \textless{}.001 & 0.032 & 0.048 \\
Non-learning & 0.107 & 0.004 & 26.415 & \textless{}.001 & 0.099 & 0.115 \\
SSL & -0.034 & 0.004 & -8.499 & \textless{}.001 & -0.042 & -0.026 \\
Hybrid Reinforce & 0.034 & 0.004 & 8.304 & \textless{}.001 & 0.026 & 0.042 \\
Participant : Trial & 0.002 & $4.11e^{-5}$ & 51.242 & \textless{}.001 & 0.002 & 0.002 \\
Habitual model : trial & -0.001 & $5.81e^{-5}$ & -21.258 & \textless{}.001 & -0.001 & -0.001 \\
MF Reinforce : Trial & -0.002 & $5.81e^{-5}$ & -26.658 & \textless{}.001 & -0.002 & -0.001 \\
Non-learning : Trial & -0.002 & $5.81e^{-5}$ & -36.189 & \textless{}.001 & -0.002 & -0.002 \\
SSL : Trial & -0.002 & $5.81e^{-5}$ & -31.573 & \textless{}.001 & -0.002 & -0.002 \\
Hybrid Reinforce : Trial & -0.002 & $5.81e^{-5}$ & -26.184 & \textless{}.001 & -0.002 & -0.001 \\ \hline
\end{tabular}
\caption{Result of the linear regression \texttt{proportion $\sim$~ model_or_participant * trial} with participant as the baseline.}
\label{table:regressionresults}
\end{table}

\bibliographystyle{apacite}

\setlength{\bibleftmargin}{.125in}
\setlength{\bibindent}{-\bibleftmargin}

\bibliography{CogSci_Template}